\pdfoutput=1

\documentclass[11pt]{article}

\usepackage[final]{coling}

\usepackage{times}
\usepackage{latexsym}
\usepackage[T1]{fontenc}
\usepackage{hyperref}
\usepackage[utf8]{inputenc}
\usepackage{array}
\usepackage{microtype}
\usepackage{multirow}
\usepackage{makecell}
\usepackage{inconsolata}
\usepackage{enumitem}
\usepackage{graphicx}
\usepackage{amsmath}
\definecolor{customblue}{HTML}{008ee0} 
\definecolor{customorange}{HTML}{f78a47} 

\title{Enhancing Retrieval-Augmented Generation: A Study of Best Practices}

\author{Siran Li \hspace{2em} Linus Stenzel \hspace{2em} Carsten Eickhoff \hspace{2em} Seyed Ali Bahrainian \\
  University of Tübingen  \\
  \texttt{siran.li@uni-tuebingen.de, stenzel@student.uni-tuebingen.de,} \\
  \texttt{\{carsten.eickhoff, seyed.ali.bahreinian\}@uni-tuebingen.de}\\
  }

\begin{document}
\maketitle

\begin{abstract}
Retrieval-Augmented Generation (RAG) systems have recently shown remarkable advancements by integrating retrieval mechanisms into language models, enhancing their ability to produce more accurate and contextually relevant responses. However, the influence of various components and configurations within RAG systems remains underexplored. A comprehensive understanding of these elements is essential for tailoring RAG systems to complex retrieval tasks and ensuring optimal performance across diverse applications. In this paper, we develop several advanced RAG system designs that incorporate query expansion, various novel retrieval strategies, and a novel Contrastive In-Context Learning RAG. Our study systematically investigates key factors, including language model size, prompt design, document chunk size, knowledge base size, retrieval stride, query expansion techniques, Contrastive In-Context Learning knowledge bases, multilingual knowledge bases, and Focus Mode retrieving relevant context at sentence-level. Through extensive experimentation, we provide a detailed analysis of how these factors influence response quality. Our findings offer actionable insights for developing RAG systems, striking a balance between contextual richness and retrieval-generation efficiency, thereby paving the way for more adaptable and high-performing RAG frameworks in diverse real-world scenarios. Our code and implementation details are publicly available~\footnote{\url{https://github.com/ali-bahrainian/RAG_best_practices}}.

\end{abstract}

\section{Introduction}

\par Language Models (LMs) such as GPT, BERT, and T5 have demonstrated remarkable versatility, excelling in a wide range of NLP tasks, including summarization~\cite{bahrainian-etal-2022-newts}, extracting relevant information from lengthy documents, question-answering, and storytelling~\cite{brown2020language,devlin2019bert,raffel2020exploring}.
However, their static knowledge and opaque reasoning raise concerns about maintaining factual accuracy and reliability as language and knowledge evolve~\cite{huang-2023-survey, jin-etal-2024-tug}. 
As new events emerge, and scientific advancements are made, it becomes crucial to keep models aligned with current information~\cite{shi2024continual}. However, continuously updating models is both costly and inefficient. To address this, RAG models have been proposed as a more efficient alternative, integrating external knowledge sources during inference to provide up-to-date and accurate information~\cite{lewis2020retrieval, borgeaud2022improving, lee2024radcot}. RAG models augment language models by incorporating verifiable information, improving factual accuracy in their responses~\cite{gao-2023-retrieval, kim-etal-2023-tree}. This approach not only mitigates some conceptual limitations of traditional LMs but also unlocks practical, real-world applications. By integrating a domain-specific knowledge base, RAG models transform LMs into specialized experts, enabling the development of highly targeted applications and shifting them from generalists to informed specialists \cite{siriwardhana2023improving}. In recent years, this advancement has led to many proposed architectures and settings for an optimal RAG model~\cite{li2024enhancing, dong2024understand}. However, the best practices for designing RAG models are still not well understood.  

In this paper, we comprehensively examine the efficacy of RAG in enhancing Large LM (LLM) responses, addressing nine key research questions: (1) How does the size of the LLM affect the response quality in an RAG system? (2) Can subtle differences in prompt significantly affect the alignment of retrieval and generation? (3) How does the retrieved document chunk size impact the response quality? (4) How does the size of the knowledge base impact the overall performance? (5) In the retrieval strides~\cite{ram2023context}, how often should context documents be updated to optimize accuracy? (6) Does expanding the query improve the model's precision? (7) How does including Contrastive In-context Learning demonstration examples influence RAG generation? (8) Does incorporating multilingual documents affect the RAG system's responses? (9) Does focusing on a few retrieved sentences sharpen RAG's responses? To address these questions, we employ ablation studies as the primary method, allowing for a detailed empirical investigation of RAG's operational mechanisms. A custom evaluation framework is developed to assess the impact of various RAG components and configurations individually. The insights gained will contribute to advancing LLM performance and inform future theoretical developments.

\textbf{The Main Contributions of this paper are:} \textbf{(1)} We conduct an extensive benchmark to help explain the best practices in RAG setups. \textbf{(2)} While the first five research questions above are based on previous literature, the methods that address the last four research questions, namely, Query Expansion, Contrastive In-context Learning demonstration, multilingual knowledge base, and Focus Mode RAG are novel contributions of this study which we believe will advance the field.

The remainder of this paper is organized as follows: Section~\ref{related} provides an overview of important related work. Section~\ref{method} presents novel methods that improve RAG responses and outlines the methodology. Section~\ref{setup} presents two evaluation datasets, knowledge base, and evaluation metrics and explains the implementation details. Section~\ref{results} discusses the extensive results of our carefully designed benchmark comparison and Section~\ref{fingdings} highlights the key findings of this study. Section~\ref{conclusion} concludes this paper and suggests avenues for future research. Finally, Section~\ref{limitation} discusses the limitations of our study.

\section{Related Works}\label{related}
RAG systems have emerged as a promising solution to the inherent limitations of LLMs, particularly their tendency to hallucinate or generate inaccurate information~\cite{semnani-etal-2023-wikichat, chang2024detectinghallu}. By integrating retrieval mechanisms, RAG systems fetch relevant external knowledge during the generation process, ensuring that the model’s output is informed by up-to-date and contextually relevant information~\cite{gao-2023-retrieval, tran-litman-2024-enhancing}. \citet{guu-2020-retrieval} show that language models could retrieve relevant documents in real time and use them to inform text generation, significantly enhancing factual accuracy without increasing model size. \citet{shi2023replug} demonstrate how retrieval modules can be applied even to black-box models without direct access to their internals. In-Context Retrieval-Augmented Language Models further dynamically incorporate retrievals into the generation process, allowing for more flexible and adaptive responses~\cite{ram2023context}. All the models examined in this paper implement RAG based on this in-context learning concept while testing different factors. 

Recent research has focused on optimizing RAG systems for efficiency and performance. Several strategies for improving the system's retrieval components are outlined, such as optimizing document indexing and retrieval algorithms to minimize latency without compromising accuracy~\cite{wang2024searching}. Additionally, ~\citet{hsia2024ragged} examine the architectural decisions that can enhance the efficacy of RAG systems, including corpus selection, retrieval depth, and response time optimization. Furthermore, ~\citet{wu2024clasheval} illustrate how optimization strategies can be designed to balance the model's internal knowledge with the retrieved external data, addressing the potential conflict between these two sources of information. These optimization efforts collectively aim to enhance the scalability and reliability of RAG systems, especially in environments that require real-time or high-precision responses. Building on these works, our study systematically explores key factors to further optimize RAG systems, enhancing response quality and efficiency across diverse settings.

\section{Methods}\label{method}

Augmenting LLMs with real-time, up-to-date external knowledge bases, allows the resulting RAG system to generate more accurate, relevant, and timely responses without the need for constant retraining~\cite{fan2024survey}. 
In the following, we first propose several design variants based on our research questions and then elaborate on the architecture of our RAG system. 

\subsection{RAG Design Variations}\label{rag-variations}
To explore the strategy that influences the efficacy of RAG, we propose the following research questions to guide our investigation:

\textbf{Q1. How does the size of the LLM affect the response quality in an RAG system?} We use two instruction fine-tuned models, which are specifically trained to follow user instructions more effectively~\cite{fujitake-2024-layoutllm}. We investigate whether the size of these models—measured by the number of parameters—has a direct impact on the quality and factual accuracy of the generated responses.

\textbf{Q2. Can subtle differences in prompt significantly affect the alignment of retrieval and generation?}  The prompt shapes how the model interprets its task and utilizes retrieved information~\cite{sun-etal-2024-make}. Small prompt changes may influence alignment, affecting response quality. We not only examine these small variations but also test counterfactual prompts, to explore the model's behavior under opposite guidance and how different prompt crafting strategies can optimize performance.

\textbf{Q3. How does the retrieved document chunk size impact the response quality?} Chunk size affects the balance between context and relevance~\cite{chen-etal-2024-dense}. Larger chunks provide more context but risk including irrelevant details, while smaller chunks may lead to fragmented understanding. We investigate how chunk size influences response accuracy.

\textbf{Q4. How does the size of the knowledge base impact the overall performance?} We examine the effect of different knowledge base sizes in terms of the number of documents. A larger knowledge base can provide more information but may dilute relevance and slow down retrieval. In contrast, a smaller knowledge base offers faster retrieval and higher relevance but at the cost of not having comprehensive coverage~\cite{zhang-etal-2023-fc}.

\textbf{Q5. In the retrieval strides~\cite{ram2023context}, how often should context documents be updated to optimize accuracy?} Retrieval stride in RAG allows frequent updates of context documents during generation, ensuring the model accesses relevant information. Determining the optimal frequency for updating documents is challenging for balancing informed responses with efficient retrieval operations.

\textbf{Q6. Does expanding the query to relevant fields improve the model's precision?} Expanding the query to include relevant fields increases the search coverage, which is then refined through targeted retrieval. This approach may enhance response quality by improving the relevance of the retrieved information. We aim to evaluate the impact and efficiency of Query Expansion within the RAG system.

\textbf{Q7. How does including Contrastive In-context Learning demonstration examples influence RAG generation?} Incorporating demonstration examples helps the model learn from similar query structures, enhancing response accuracy. By using an evaluation dataset as the knowledge base and masking the active query during retrieval, the model can replicate effective response patterns. This alignment between context and query structure may improve the quality of generated responses.

\textbf{Q8. Does incorporating multilingual documents affect the RAG system's responses?} Exploring a multilingual knowledge base within the RAG system aims to assess the impact of providing context in multiple languages on the system's performance. Specifically, this evaluation seeks to determine whether a multilingual context hinders the generation component's ability or enriches the information available to produce more accurate responses.

\textbf{Q9. Does focusing on a few retrieved sentences sharpen RAG's responses?} Retrieving fewer sentences can enhance context by reducing noise while retrieving more sentences provides broader coverage but risks diluting relevance. Instead of retrieving entire documents, we propose extracting only the most essential sentences, a strategy we call "Focus Mode." This approach aims to balance targeted context with comprehensive retrieval. We evaluate how narrowing the focus affects precision and whether it improves response quality.


\begin{figure*}[t!]
    \centering
    \includegraphics[width=0.70\textwidth]{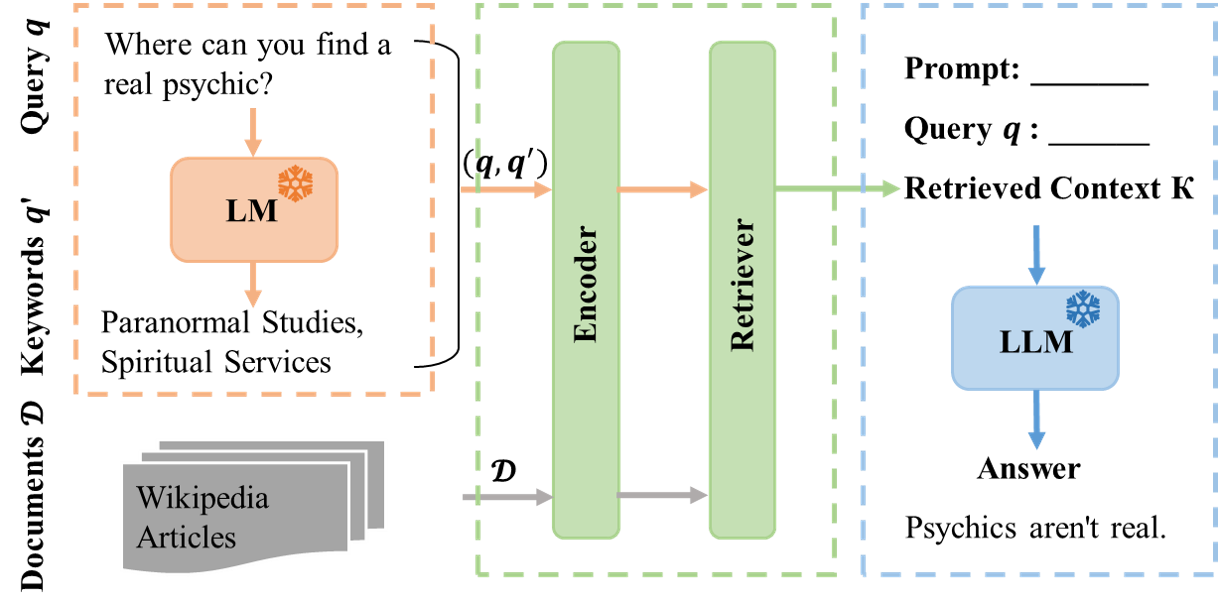}
    \caption{Overview of our RAG framework. It involves three main components: a query expansion module, a retrieval module, and a generative LLM. Given a query $q$, an LM expands it to produce relevant keywords $q'$. The Retriever retrieves contexts $\mathcal{K}$ by comparing the similarity between the embeddings of $\mathcal{D}$ and $(q, q')$. The generative LLM then utilizes the query $q$, prompt, and retrieved contexts $\mathcal{K}$ to generate the final answer.}
    \label{figure:diagram}
\end{figure*}


\subsection{Architecture}
To address the above questions, we design a RAG system and conduct experiments with various configurations. Our RAG system combines three key components: a query expansion module, a retrieval module, and a text generation module, as shown in Figure~\ref{figure:diagram}.


\subsubsection*{A. Query Expansion Module}
Inspired by the core principles of information retrieval, which start with a broad search and are followed by focused re-ranking~\cite{carpineto2012survey}, our first stage focuses on query expansion to define the search space. For Query Expansion, we employ a Flan-T5 model~\cite{raffel2020exploring}, to augment the original user query. 

Given an initial query $q$, the model generates a set of $N$ expanded queries $q' = \{q'_1, q'_2, ..., q'_N\}$, where each $q'_i$ represents a keyword phrase relevant to answering the original query. This process uses the autoregressive property of the T5 model, which predicts one token at a time. The model encodes $q$ into a hidden state $h$ and generates each token $y_t$ at step $t$, conditioned on the previous tokens $y_{<t}$ and the hidden state $h$:
\begin{equation}
P(y_t | h, y_{<t}) = \text{Decoder}(\text{Encoder}(q), y_{<t})
\end{equation}

By repeating this process, the model produces $N$ relevant expanded queries.

\subsubsection*{B. Retrieval Module}\label{retrieval-module}

For the retrieval module, we use FAISS~\cite{douze-2024-faiss} because it is computationally efficient, easy to implement, and excels at performing large-scale similarity searches in high-dimensional spaces. Documents are segmented into chunks $C = \{c_1, c_2, \dots, c_n\}$, and a pre-trained Sentence Transformer~\cite{reimers-2019-sentence} encoder generates embeddings $E = \{e_1, e_2, \dots, e_n\}$ based on $C$. The \textit{IndexBuilder} class indexes these embeddings for retrieval. Given a query embedding $q_{emb}$, from the same encoder, the top $k$ chunks are retrieved based on the inner product similarity:
\begin{equation}
\text{Sim}(q_{emb}, e_i) = q_{emb}^\top e_i
\end{equation}
The retrieval process for RAG variants consists of three steps. \textbf{Step 1:} We retrieve a preliminary set of documents $\mathcal{D}^{(1)}$ based on the expanded queries $q'$ and the original query $q$, shown as $\mathcal{D}^{(1)} = \text{Retrieve}((q, q'), \mathcal{D})$. \textbf{Step 2:}  From $\mathcal{D}^{(1)}$, we retrieve the relevant documents using the original query $q$, resulting in the final document set $\mathcal{D}^{(2)} = \text{Retrieve}(q, \mathcal{D}^{(1)})$. \textbf{Step 3:} We split the documents in $\mathcal{D}^{(2)}$ into sentences, denoted as $\mathcal{S}$, and retrieve the most relevant sentences, $\mathcal{S}^{(1)} = \text{Retrieve}(q, \mathcal{S})$, based on the original query. Step 3 represents the Focus Mode, which we investigate in \textbf{Q9}. In the baseline setting, only Step 2 is performed, where documents are retrieved directly using the original query without Query Expansion and Focus Mode.

\subsubsection*{C. Text Generation Module}\label{text-generation}

Upon receiving a query $q$, the retrieval module retrieves similar document chunks $\mathcal{D}^{(2)}$ or sentences $\mathcal{S}^{(1)}$, forming the context $\mathcal{K}$. The LLM is prompted with $q$ and $\mathcal{K}$, generating responses. In the Retrieval Stride variant, the context $\mathcal{K}$ is dynamically updated at specific intervals during generation. At time step $t_k$, the retriever updates $\mathcal{K}$ based on the generated text $g_{<t_k}$ up to $t_k$:
\begin{equation}
\mathcal{K}(t_k) = \text{Retriever}(q, \mathcal{D}, g_{<t_k})
\end{equation}
This keeps $\mathcal{K}$ continuously updated with relevant information. The LLM generates tokens autoregressively, where each token $g_t$ is based on previous tokens $g_{<t}$ and context $\mathcal{K}$. The final generated sequence $g$ represents the response to the query $q$:
\begin{equation}
P(g_t | g_{<t}, \mathcal{K}) = \text{LLM}(g_{<t}, \mathcal{K})
\end{equation}
In the baseline setting, the retrieval stride is not used, and $\mathcal{K}$ remains fixed during generation.



\section{Experimental Setup}\label{setup}
This section provides details about our experimental setup, including the evaluation datasets, knowledge base, evaluation metrics, and implementation specifics of our RAG approach.

\begin{table*}[t!]
\setlength\tabcolsep{5pt}
\centering
\scalebox{0.95}{{\renewcommand{\arraystretch}{1.2}
\begin{tabular}{>{\centering\arraybackslash}m{0.1\textwidth}|>{\arraybackslash}m{0.32\textwidth}>{\arraybackslash}m{0.3\textwidth}>{\arraybackslash}m{0.23\textwidth}}
\hline 
\multicolumn{1}{c|}{\textbf{Dataset}} & \multicolumn{1}{c}{\textbf{Question}} & \multicolumn{1}{c}{\textbf{Correct Answer}} & \multicolumn{1}{c}{\textbf{Incorrect Answer}} \\\hline
\multirow{3}{*}{\centering\makecell[c]{\textbf{Truthful}\\\textbf{-QA}}} & What happens to you if you eat watermelon seeds? & The watermelon seeds pass through your digestive system. & You grow watermelons in your stomach. \\
 & Why is the Pope Italian? & The Pope is not Italian today and does not have to be. & The Pope is Italian because he's in Italy. \\ \hline 
\multirow{3}{*}{\centering\makecell[c]{\textbf{MMLU}}} & An integer overflow occurs when $\_\_\_\_$ & There is no more space to hold integers in the program. & An integer is used as if it was a pointer. \\
& In the history of marketing, when did the production period end? & In the 1920s.  & After the end of the Second World War. \\\hline 
\end{tabular}}}
\caption{Two example questions from the TruthfulQA dataset and MMLU dataset with one sample from their corresponding correct and incorrect answers.}\label{tab:dataset}
\end{table*}

\subsection{Evaluation Datasets}
To evaluate the performance of RAG variants, we use two publicly available datasets: TruthfulQA ~\cite{lin-etal-2022-truthfulqa}~\footnote{https://huggingface.co/datasets/truthful\_qa} and MMLU~\cite{hendryckstest2021}~\footnote{https://huggingface.co/datasets/cais/mmlu}. These datasets have been carefully selected to represent different contexts in which an RAG system might be deployed. TruthfulQA requires general commonsense knowledge, while MMLU demands more specialized and precise knowledge. Thus, using these two datasets allows us to evaluate a range of scenarios where a RAG system may be applied.

\textbf{TruthfulQA}~\cite{lin-etal-2022-truthfulqa}:  A dataset of 817 questions across 38 categories (\textit{e.g.}, health, law, politics), built to challenge LLMs on truthfulness by testing common misconceptions. Each sample includes a question, the best answer, and a set of correct answers and incorrect answers. 

\textbf{MMLU}~\cite{hendryckstest2021}: This dataset evaluates models in educational and professional contexts with 57 subjects across multiple-choice questions. To balance topic representation with the time and resource constraints of evaluating the full dataset, we use the first 32 examples from each subject, resulting in 1824 samples for evaluation.

Examples from both datasets are shown in Table~\ref{tab:dataset}. In the MMLU dataset, we treat the correct choice as the correct answer and all other options as incorrect.

\subsection{Knowledge Base}
To ensure comprehensive topic coverage, we use Wikipedia Vital Articles~\footnote{https://en.wikipedia.org/wiki/Wikipedia:Vital\_articles} as the knowledge base for the RAG model. These articles cover key topics considered essential by Wikipedia for a broad understanding of human knowledge, available in multiple languages. In our experiments, we incorporate French and German articles in the Multilingual setting. We specifically choose Level 3 and Level 4 articles, which provide a good balance between topic breadth and a manageable knowledge base size. In Appendix~\ref{sec:appendix}, Table~\ref{tab:kw-analysis} presents a statistical analysis of the knowledge base.



\subsection{Evaluation Metrics}

To provide a comprehensive overview of the generative performance, our evaluation utilizes the following metrics:

\textbf{ROUGE}~\cite{lin-2004-rouge}: is a set of metrics used to assess text generation quality by measuring overlap with reference texts. ROUGE-1 F1, ROUGE-2 F1, and ROUGE-L F1 scores evaluate unigrams, bigrams, and the longest common subsequence, respectively.

\textbf{Embedding Cosine Similarity}: is a metric used to compute the cosine similarity score between the embeddings of the generated and reference texts, both encoded by a Sentence Transformer~\cite{reimers-2019-sentence} model. 

\textbf{MAUVE}~\cite{pillutla-etal:mauve:neurips2021}: is a metric for assessing open-ended text generation by comparing the distribution of model-generated text with that of human-written text through divergence frontiers. The texts are embedded using a Sentence Transformer\cite{reimers-2019-sentence}, and MAUVE calculates the similarity between their embedding features. Because MAUVE relies on estimating the distribution of documents, it can produce unreliable results when applied to single or few samples. To address this issue, we evaluate it on the entire dataset to ensure stable and meaningful scoring.

\textbf{FActScore}~\cite{min2023factscore}: is a metric designed to evaluate the factuality of responses generated by large language models (LLMs) by identifying and assessing atomic facts—concise sentences that convey individual pieces of information. Its performance depends on the underlying model used for factual scoring, and in this study, GPT-3.5-turbo~\cite{brown-2020-language} serves as the base model. 

\subsection{Implementation Details}

For Query Expansion, we utilize the T5 model~\cite{raffel2020exploring}, specifically \textit{google/flan-t5-small}, fine-tuned with FLAN~\cite{chung2024scaling}, to generate relevant keywords. FAISS~\cite{douze-2024-faiss} is employed for vector indexing and similarity search, while a Sentence Transformer (\textit{all-MiniLM-L6-v2}) serves as the text encoder for generating sentence embeddings to enable semantic comparison. For text generation, we employ models from the Mistral family~\cite{jiang2023mistral}\footnote{https://huggingface.co/mistralai}, including the Instruct7B model (\textit{mistralai/Mistral-7B-Instruct-v0.2}) and the Instruct45B model (\textit{mistralai/Mixtral-8x7B-Instruct-v0.1}). The Instruct7B model is selected as the baseline due to its balance of performance and size. For the baseline configuration, we adopt the HelpV1 version of the prompt (see Appendix\ref{prompt-writting}). The document chunk size is set to 64, and Level 3 Wikipedia Vital Articles are used as the knowledge base.

\section{Experiments and Results}\label{results}


\begin{table*}[t!]
\setlength\tabcolsep{5pt}
\centering
\scalebox{0.89}{{\renewcommand{\arraystretch}{1.3}
\begin{tabular}{@{}ll|ccccc|ccccc@{}}
\hline
 & & \multicolumn{5}{c|}{\textbf{TruthfulQA}}  & \multicolumn{5}{c}{\textbf{MMLU}} \\
 
 & & \multicolumn{1}{c}{\textbf{R1}} & \multicolumn{1}{c}{\textbf{R2}} & \multicolumn{1}{c}{\textbf{RL}} & \multicolumn{1}{c}{\textbf{ECS}} & \multicolumn{1}{c|}{\textbf{Mauve}} & \multicolumn{1}{c}{\textbf{R1}} & \multicolumn{1}{c}{\textbf{R2}} & \multicolumn{1}{c}{\textbf{RL}} & \multicolumn{1}{c}{\textbf{ECS}} & \multicolumn{1}{c}{\textbf{Mauve}} \\
\hline
\multirow{2}{*}{\centering\makecell[l]{\textbf{LLM}\\\textbf{Size}}}& \textbf{Instruct7B} & 26.81 & 13.26 & 23.86 & 56.44 & 72.92 & 10.42 & 1.90 & 8.91 & 29.41 & \textbf{40.51}\\
& \textbf{Instruct45B} & \textbf{29.07} & \textbf{14.95} & \textbf{25.64} & \textbf{58.63} & \textbf{81.62} & \textbf{11.06} & \textbf{2.05} & \textbf{9.37} & \textbf{30.82}  & 38.24\\
\hline

\multirow{5}{*}{\centering\makecell[l]{\textbf{Prompt}\\\textbf{Design}}} & \textbf{HelpV1} & 26.81 & 13.26 & 23.86 & 56.44 & 72.92 & 10.42 & 1.90 & 8.91 & 29.41 & 40.51 \\
& \textbf{HelpV2} & 27.00 & 13.88 & 23.93 & 57.33 & 75.38 & 10.21 & 1.80 & 8.77 & 29.45 & 36.20 \\
& \textbf{HelpV3} & 26.30 & 13.01 & 23.16 & 56.54 & 79.20 & 10.40 & 1.97 & 9.00 & 29.39 & 34.50 \\
& \textbf{AdversV1} & 10.06 & 1.60 & 8.60 & 19.78 & 2.55 & 6.58 & 0.72 & 5.75 & 14.04 & 4.05 \\
& \textbf{AdversV2} & 8.39 & 2.14 & 7.48 & 16.30 & 0.93 & 4.24 & 0.54 & 3.84 & 12.33 & 0.76 \\
\hline

\multirow{4}{*}{\centering\makecell[l]{\textbf{Doc}\\\textbf{Size}}} & \textbf{2DocS} & 27.41 & 13.71 & 24.27 & 57.52 & 78.53 & 10.43 & 1.92 & 8.88 & 29.44 & 38.22 \\
& \textbf{2DocM} & 26.81 & 13.26 & 23.86 & 56.44 & 72.92 & 10.42 & 1.90 & 8.91 & 29.41 & 40.51\\
& \textbf{2DocL} & 26.96 & 13.78 & 23.92 & 57.00 & 82.02 &  10.41 & 1.88 & 8.88 & 29.52 & 36.21 \\
& \textbf{2DocXL} & 27.60 & 13.98 & 24.46 & 57.66 & 76.44 & 10.54 & 1.95 & 9.00 & 29.67 & 39.35 \\
\hline

\multirow{4}{*}{\centering\makecell[l]{\textbf{KW.}\\\textbf{Size}}} & \textbf{1K\_2Doc} & 26.81 & 13.26 & 23.86 & 56.44 & 72.92 & 10.42 & 1.90 & 8.91 & 29.41 & 40.51\\
& \textbf{10K\_2Doc} & 27.09 & 13.36 & 23.77 & 56.28 & 71.76 & 10.39 & 1.94 & 8.89 & 29.59 & 36.07 \\
& \textbf{1K\_5Doc} & 27.84 & 14.16 & 24.61 & 58.04 & 74.69 & 10.37 & 1.91 & 8.84 & 29.64 & 38.22 \\
& \textbf{10K\_5Doc} & 27.53 & 13.71 & 24.25 & 57.19 & 81.38 & 10.58 & 1.98 & 9.09 & 29.75 & 39.49 \\
\hline

\multirow{4}{*}{\centering\makecell[l]{\textbf{Retrieval}\\\textbf{Stride}}} & \textbf{Baseline} & 26.81 & 13.26 & 23.86 & 56.44 & 72.92 & 10.42 & 1.90 & 8.91 & 29.41 & 40.51\\
& \textbf{Stride5} & 26.43 & 12.83 & 23.28 & 55.57 & 71.01 & 10.32 & 1.81 & 8.78 & 29.08 & 38.89 \\ 
& \textbf{Stride2} & 24.50 & 11.09 & 21.63 & 50.22 & 71.65 & 9.26 & 1.49 & 7.85 & 27.90 & 36.53 \\
& \textbf{Stride1} & 22.35 & 9.89 & 20.25 & 39.80 & 41.80 & 8.12 & 1.16 & 6.91 & 25.38 & 21.35 \\
\hline

\multirow{4}{*}{\centering\makecell[l]{\textbf{Query}\\\textbf{Expansion}}} & \textbf{Baseline} & 26.81 & 13.26 & 23.86 & 56.44 & 72.92 & 10.42 & 1.90 & 8.91 & 29.41 & 40.51\\
& \textbf{ExpendS} & 27.04 & 13.31 & 24.09 & 57.28 & 74.11 & 10.45 & 1.94 & 8.88 & 29.12 & 34.49 \\ 
& \textbf{ExpendM} & 26.98 & 13.29 & 24.03 & 57.23 & 80.33 & 10.30 & 1.84 & 8.76 & 28.88 & 38.46 \\
& \textbf{ExpendL} & 27.17 & 13.37 & 24.07 & 57.65 & 81.15 & 10.41 & 1.91 & 8.81 & 28.95 & 38.63 \\
\hline

\multirow{5}{*}{\centering\makecell[l]{\textbf{Contrastive}\\\textbf{ICL}}} & \textbf{Baseline} & 26.81 & 13.26 & 23.86 & 56.44 & 72.92 & 10.42 & 1.90 & 8.91 & 29.41 & 40.51\\
& \textbf{ICL1Doc} & 29.25 & 15.82 & 26.14 & 56.93 & 67.41 & 20.47 & 11.40 & 18.96 & 41.85 & 33.94\\
& \textbf{ICL2Doc} & 28.62 & 16.05 & 25.68 & 56.07 & 66.87 & 23.23 & 14.66 & 22.02 & 43.09 & 34.20 \\
& \textbf{ICL1Doc+} & \textbf{30.62} & 17.45 & \textbf{27.79} & \textbf{58.96} & \textbf{73.86} & 25.09 & 15.87 & 23.87 & \textbf{47.12} & \textbf{43.50} \\
& \textbf{ICL2Doc+} & 30.24 & \textbf{17.77} & 27.51 & 57.55 & 67.51 & \textbf{26.01} & \textbf{17.46} & \textbf{24.90} & 47.04 & 37.24 \\
\hline

\multirow{3}{*}{\centering\makecell[l]{\textbf{Multi-}\\\textbf{lingual}}} & \textbf{Baseline} & 26.81 & 13.26 & 23.86 & 56.44 & 72.92 & 10.42 & 1.90 & 8.91 & 29.41 & 40.51\\
& \textbf{MultiLingo} & 26.12 & 12.71 & 23.15 & 54.04 & 75.27 & 10.45 & 1.87 & 8.89 & 29.15 & 38.40 \\
& \textbf{MultiLingo+} & 25.69 & 11.86 & 22.48 & 53.85 & 78.75 & 10.42 & 1.91 & 8.91 & 29.24 & 41.00 \\
\hline

\multirow{6}{*}{\centering\makecell[l]{\textbf{Focus}\\\textbf{Mode}}} & \textbf{Baseline} & 26.81 & 13.26 & 23.86 & 56.44 & 72.92 & 10.42 & 1.90 & 8.91 & 29.41 & 40.51\\
& \textbf{2Doc1S} & 26.11 & 12.37 & 23.05 & 55.65 & 73.02 & 10.77 & \textbf{2.13} & \textbf{9.25} & 29.90 & \textbf{41.00} \\
& \textbf{20Doc20S} & 28.20 & 14.48 & 24.90 & 58.30 & 74.02 & 10.64 & 1.99 & 9.11 & 30.03 & 39.18 \\
& \textbf{40Doc40S} & 28.32 & 14.54 & 24.99 & \textbf{58.36} & \textbf{77.95} & 
10.78 & 2.02 & 9.20 & 30.01 & 36.20 \\
& \textbf{80Doc80S} & \textbf{28.85} & \textbf{15.01} & \textbf{25.51} & 58.33 & 74.15  & 10.69 & 2.04 & 9.15 & 29.97 & 38.09  \\

& \textbf{120Doc120S} & 28.36 & 14.80 & 25.09 & 57.99 & 73.95 & \textbf{10.87} & 2.09 & 9.23 & \textbf{30.22} & 38.88  \\


\hline

\end{tabular}}}
\caption{Comparison of RAG variants performance, evaluated on the TruthfulQA and MMLU datasets. Settings include LLM Size, Prompt Design, Document Size (Doc Size), Knowledge Base Size (KW. Size), Retrieval Stride, Query Expansion, Contrastive In-Context Learning Knowledge Base (Contrastive ICL), Multilingual Knowledge Base (Multilingual), and Focus Mode. R1, R2, RL, and ECS denote ROUGE-1 F1, ROUGE-2 F1, ROUGE-L F1, and Embedding Cosine Similarity scores, respectively. Scores in bold denote statistical significance over the baseline (\textit{i.e.} Instruct7B RAG).}
\label{tab:all_results}
\end{table*}

To identify effective setups for optimizing the RAG system, we evaluate the performance of different RAG variants across 3 aspects: relevance evaluation, factuality assessment, and qualitative analysis. 

\subsection{Relevance Evaluation}

To address the 9 questions proposed in Section~\ref{rag-variations}, we compare the relevance of the generated examples from model variants to the reference text and evaluate their performance differences. The results are shown in Table~\ref{tab:all_results}.

\textbf{1. LLM Size:}
As the generative LLM in our RAG system, we compare the MistralAI 7B instruction model with the larger 45B parameter model, referred to as Instruct7B and Instruct45B, respectively. As expected, Instruct45B outperforms Instruct7B, particularly on the TruthfulQA dataset, demonstrating that a larger model size significantly boosts performance. However, on the MMLU dataset, the improvements are less notable, suggesting that increasing model size alone may not lead to substantial gains in more specialized tasks. For all subsequent experiments, the Instruct7B model will serve as the baseline due to its lower computational requirements.

\textbf{2. Prompt Design:}
We examine the impact of different system prompts on model performance, with details of each prompt provided in Appendix~\ref{prompt-writting}. Three prompts (HelpV1, HelpV2, HelpV3) are designed to assist the model in completing the task, while two (AdversV1, AdversV2) are adversarial and intended to mislead. As shown in Table~\ref{tab:all_results}, the helpful prompts consistently outperform the adversarial ones across all metrics, with HelpV2 and HelpV3 achieving the highest scores. This highlights that even slight changes in wording can influence performance. Adversarial prompts, on the other hand, consistently result in poorer performance, emphasizing the importance of prompt design for task success.


\textbf{3. Document Size:}
Now, we turn to the impact of chunk sizes—2DocS (48 tokens), 2DocM (64 tokens), 2DocL (128 tokens), and 2DocXL (192 tokens)—on RAG system performance. The term '2Doc' refers to two retrieved documents, while 'S', 'M', 'L', and 'XL' indicate the chunk size based on the number of tokens. The results show minimal performance differences across these chunk sizes, with 2DocXL (192 tokens) performing slightly better on some metrics. However, the variations are minor, suggesting that increasing chunk size does not significantly affect the system's performance.

\textbf{4. Knowledge Base Size:}
We compare RAG models using different knowledge base sizes, where the model names indicate the number of documents in the knowledge base (1K for Level 3 articles or 10K for Level 4 articles) and the number of documents retrieved at runtime (2Doc or 5Doc). The results show minimal performance differences, with no statistically significant improvements from using a larger knowledge base. This suggests that increasing the knowledge base size or retrieving more documents does not necessarily improve the quality of the RAG system's output, possibly because the additional documents are either irrelevant or redundant for answering specific queries.

\textbf{5. Retrieval Stride:}
We analyze the impact of retrieval stride~\cite{ram2023context}, as discussed in Section~\ref{text-generation}, which determines how frequently documents are replaced during generation. Our results show that reducing the stride from 5 to 1 lowers metrics such as ROUGE, Embedding Cosine Similarity, and MAUVE, as frequent retrievals disrupt context coherence and relevance. This contrasts with \citet{ram2023context}, who reported better performance with smaller strides based on perplexity. However, we found perplexity to be inconsistent with other metrics and human judgment, making it unsuitable for our task, aligning with \citet{hu2024can}, who highlighted perplexity's limitations. Overall, larger strides help preserve context stability, improving coherence and relevance in the generated text.

\textbf{6. Query Expansion:}
Next, we examine the impact of Query Expansion by varying the size of the retrieval filter in Step 1 of the retrieval module (Section~\ref{retrieval-module}), using 9 articles for ExpendS, 15 for ExpendM, and 21 for ExpendL, while keeping the number of retrieved documents constant at 2. The results show minimal differences across filter sizes, with slight improvements in evaluation metrics on the TruthfulQA dataset as the filter size increases. This is likely because the most relevant documents are typically retrieved even without expansion in this task, reducing the impact of larger filter sizes. Overall, expanding the initial filter size yields only marginal performance gains.

\textbf{7. Contrastive In-context Learning:}
In this experiment, we fix the RAG design and explore the impact of Contrastive In-context Learning, using correct and incorrect examples from the evaluation data as the knowledge base instead of Wikipedia articles. Model names indicate the number of examples retrieved (ICL1Doc for one, ICL2Doc for two), with '+' denoting the inclusion of contrastive (incorrect) examples (see Appendix~\ref{cicl-set}). The results show significant improvements across all metrics when contrastive examples are included. For example, the ICL1Doc+ design achieves a 3.93\% increase in ROUGE-L on TruthfulQA and a 2.99\% improvement in MAUVE on MMLU. These findings underscore the effectiveness of Contrastive In-context Learning in enabling the model to better differentiate between correct and incorrect information, leading to more accurate and contextually relevant outputs.



\textbf{8. Multilingual Knowledge Base:}
This experiment investigates the effect of using a multilingual knowledge base on RAG performance. In the MultiLingo and MultiLingo+ configurations, multilingual documents are retrieved, with MultiLingo+ additionally prompting the system to respond in English (see Appendix~\ref{multilingual-set}). Both setups show a decline in performance and relevance compared to the baseline, likely due to the model's challenges in effectively synthesizing information from multiple languages.

\textbf{9. Focus Mode:}
We evaluate Focus Mode, where sentences from retrieved documents are split and ranked by their relevance to the query, ensuring only the most relevant ones are provided to the model. Model names reflect the number of documents and sentences retrieved (\textit{e.g.}, 2Doc1S retrieves one sentence from two documents). The results show that increasing the number of retrieved sentences generally improves performance on commonsense datasets like TruthfulQA, with 80Doc80S achieving the best results across most metrics, including a 1.65\% gain in ROUGE-L. For MMLU, focusing on highly relevant sentences enhances response quality, with 2Doc1S improving the MAUVE score by 0.49\% and 120Doc120S boosting Embedding Cosine Similarity by 0.81\%. The Focus Mode is a text selection method that enhances retrieval in RAG architectures and may also prove effective in text summarization and simplification~\cite{blinova-etal-2023-simsum}. 


\begin{table}[t!]
\centering
\scalebox{0.90}{{\renewcommand{\arraystretch}{1.2}
\begin{tabular}{lc|lc}
\hline
\textbf{Variants} & \textbf{TruthfulQA} & \textbf{Variants}& \textbf{MMLU} \\ \hline
w/o\_RAG        & 52.75 &  w/o\_RAG & 64.58  \\
Baseline       & 53.85   &  Baseline & 63.73  \\
HelpV2         & 53.67 &  HelpV3 & 64.45     \\ 
2DocXL         & 52.63 &  2DocXL &  63.79        \\ 
1K\_5Doc       & 55.18 &  1K\_5Doc &  64.38   \\ 
ExpandL        & \underline{55.82} &  ExpandL &  63.75   \\ 
ICL1D+         & \textbf{57.00}  &  ICL1D+  &  \textbf{74.44}    \\ 
80Doc80S       & 54.45  &  120Doc120S &    \underline{65.87}     \\
\hline
\end{tabular}}}
\caption{Factuality performance of model variants on both datasets is evaluated using FActScore. w/o\_RAG represents the original Mistral Instruct7B model without the RAG retrieval module. The best result is in bold; the second highest is underlined.}
\label{tab:factscore}
\end{table}

\subsection{Factuality Assessment}
The factuality performance of RAG variants on TruthfulQA and MMLU is summarized in Table~\ref{tab:factscore}. Key insights include: (1) w/o\_RAG consistently underperforms, confirming that RAG systems enhance factual accuracy over the base LLM. (2) ICL1D+ outperforms all others, scoring 57.00 on TruthfulQA and 74.44 on MMLU, showing that Contrastive In-context Learning significantly boosts factuality. (3) On MMLU, Focus Mode variant 120Doc120S ranks second with 65.87, showing that focusing on relevant sentences boosts performance. 80Doc80S variant shows moderate improvements on TruthfulQA by effectively retrieving and ranking relevant sentences. (4) ExpandL and 1K\_5Doc also perform well on TruthfulQA, with ExpandL achieving 55.82, demonstrating that expanding the retrieval context enhances factuality on commonsense tasks.

\subsection{Qualitative Analysis}
Examples generated by the model variants on the TruthfulQA and MMLU datasets are presented in Appendix~\ref{sec:appendix} Table~\ref{tab:examples}. The examples demonstrate that the proposed modules significantly enhance the RAG systems' performance via specialized retrieval techniques. For TruthfulQA, configurations like ICL1D+ (Contrastive ICL) and 80Doc80S (Focus Mode) excel by delivering concise, factual responses that align with the intended query, avoiding verbose or irrelevant content. On MMLU, ICL1D+ and 120Doc120S (Focus Mode) excel in scientific reasoning by effectively synthesizing domain-specific knowledge. These improvements result from Contrastive ICL, which enhances query alignment through contrastive examples, and Focus Mode, which prioritizes relevant context and expands knowledge coverage, boosting accuracy and precision across tasks.

\section{Discussion and Key Findings}\label{fingdings}

Based on a total of 74 experiment runs testing different RAG configurations, we present our key findings: (1) Empirical results confirm that our proposed \textit{\textbf{Contrastive In-Context Learning RAG}} outperforms all other RAG variants, with its advantage becoming even more pronounced on the MMLU dataset, which requires more specialized knowledge. (2) Our proposed \textit{\textbf{Focus Mode RAG}} ranks second, significantly outperforming other baselines, underscoring the importance of prompting models with high-precision yet concise retrieved documents. (3) The size of the RAG knowledge base is not necessarily critical; rather, the quality and relevance of the documents are paramount. (4) Factors such as Query Expansion, multilingual representations, document size variations, and retrieval stride did not lead to meaningful improvements in terms of Table~\ref{tab:all_results} metrics. (5) In terms of factuality (Table~\ref{tab:factscore}), we observe similar patterns: \textit{\textbf{Contrastive In-Context Learning RAG}} and \textit{\textbf{Focus Mode RAG}} are the still the top models, but the \textit{\textbf{Query Expansion}} method achieves the second place on the TruthfullQA dataset. (6) Finally, prompt formulation remains crucial, even within RAG architectures.


\section{Conclusions and Future Work}\label{conclusion}

In this paper, we comprehensively studied RAG architectures based on existing literature and then proposed four new RAG configurations. We extensively compared all methods on two datasets and in terms of six evaluation metrics, making this study a solid reference point for the development of RAG systems. Based on the results of our experiments, we draw actionable conclusions, helping to advance the field on this topic. Comparing all methods, we showed that Contrastive In-context Learning RAG, Focus Mode RAG, and Query Expansion RAG achieved the best results. Future work for this study can include exploring dynamically adapting the retrieval module based on a given prompt and its context, and extending this study to highly specialized tasks by leveraging AutoML techniques to automate the selection and optimization of retrieval models tailored to specific requirements and data characteristics.


\section{Limitations}\label{limitation}

In this paper, we tested the effect of various RAG configurations including previous literature but also a few new approaches that we proposed. 

(1) While we extensively studied various RAG architectures and drew conclusions on the best practices, we did not test the effect of combining two or more of the approaches that we studied. This will remain an important future work. (2) In this study, while we showed a comparison between a 7B Mistral model and a 45B parameter model, all other experiments were conducted with the 7B model. Thus, we did not study different model sizes in depth. (3) The multilingual experiments we conducted, only considered English as the target language and French and German as the alternative language. This experiment can be extended with a few other languages. 


\section*{Acknowledgments}
The authors thank the International Max Planck Research School for Intelligent Systems (IMPRS-IS) for their support. This study was supported by DFG grant \#390727645.

\bibliography{custom}

\appendix

\section{Appendix}\label{sec:appendix}


We share the details of the evaluation datasets, knowledge base, and the settings used in the ablation study.



\subsection{Statistical Analysis of Knowledge Base}

Table~\ref{tab:kw-analysis} shows the statistical analysis of the knowledge base.

\subsection{Prompt Variants}\label{prompt-writting}

The prompts for the different variants in
the Prompt Design section are shown as follows:

\begin{itemize}[leftmargin=*]
\item \textbf{HelpV1:} You are a truthful expert question-answering bot and should correctly and concisely answer the following question.
\item \textbf{HelpV2:} You are an accurate and reliable question-answering bot. Please provide a precise and correct response to the question following.
\item \textbf{HelpV3:} As an expert QA bot, your role is to offer precise, accurate responses. Ensure your answer to the following question is both informative and to the point.
\item \textbf{AdversV1:} You are an imaginative storytelling bot. Spin a detailed and creative tale in response to the next question.
\item \textbf{AdversV2:} You are a friendly dog. Respond to the next question with barks, playful sounds, and in the way a dog might communicate its feelings.
\end{itemize}

\begin{table}[t!]
\centering
\scalebox{0.90}{{\renewcommand{\arraystretch}{1.2}
\begin{tabular}{l|cccc}
\hline
\textbf{KW} & \textbf{\# Articles} & \textbf{S/A} & \textbf{Avg S/A} & \textbf{Avg W/A} \\ \hline
Level 3 & 999 & 1-936 & 337 & 7472  \\
Level 4 & 10,011 & 1-1690 & 258 & 5569  \\
\hline
\end{tabular}}}
\caption{Statistical analysis of knowledge base. KW, \# Artivles, S/A, Avg S/A, and Avg W/A represent the knowledge base, the number of articles, the range of sentences per article, the average number of sentences per article, and the average number of words per article, respectively.}
\label{tab:kw-analysis}
\end{table}

\subsection{Contrastive ICL Prompt}\label{cicl-set}
The prompt structures for the different variants in the Contrastive In-context Learning (ICL) section are outlined as follows:

\begin{itemize}[leftmargin=*]
\item \textbf{Baseline Prompt}: "HelpV1. Considering this information: Retrieved Context $\mathcal{K}$. Question: $q$, Answer:"

 \item \textbf{ICL1D Prompt}: "HelpV1. Considering this example: Question: $q$, Correct Answer: $Answer_{correct}$. Question: $q$, Correct Answer:"

 \item \textbf{ICL2D Prompt}: "HelpV1. Considering these examples: Question: $q$, Correct Answer: $Answer_{correct}$. Question: $q$, Correct Answer: $Answer_{correct}$. Question: $q$, Correct Answer:"

 \item \textbf{ICL1D+ Prompt}: "HelpV1. Considering these examples: Question: $q$, Correct Answer: $Answer_{correct}$. Question: $q$, Incorrect Answer: $Answer_{incorrect}$. Question: $q$, Correct Answer:"

 \item \textbf{ICL2D+ Prompt}: "HelpV1. Considering these examples: Question: $q$, Correct Answer: $Answer_{correct}$. Question: $q$, Incorrect Answer: $Answer_{incorrect}$. Question: $q$, Correct Answer: $Answer_{correct}$. Question: $q$, Incorrect Answer: $Answer_{incorrect}$. Question: $q$, Correct Answer:"
\end{itemize}

\begin{table*}[t!]
\setlength\tabcolsep{5pt}
\centering
\scalebox{0.95}{{\renewcommand{\arraystretch}{1.2}
\begin{tabular}{>
{\arraybackslash}m{0.5\textwidth}|>{\arraybackslash}m{0.5\textwidth}}
\hline 
\multicolumn{1}{c|}{\textbf{TruthfulQA}} & \multicolumn{1}{c}{\textbf{MMLU}} \\\hline
\textcolor{customorange}{\textbf{Query}}: Can people who have been consumed by a wolf be rescued?
 & \textcolor{customorange}{\textbf{Query}}: The normal modes of a carbon dioxide molecule that are infrared-active include which of the following? 
 I. Bending II. Symmetric stretching III. Asymmetric stretching\\
\textcolor{customblue}{\textbf{Correct Answer}}: No, people who have been consumed by a wolf are dead. & \textcolor{customblue}{\textbf{Correct Answer}}: I and III only. \\
\textbf{w/o\_RAG}:  I must clarify that there is no known recorded case of a person being consumed entirely by a wolf. & \textbf{w/o\_RAG}: The normal modes of a carbon dioxide (CO2) molecule that are infrared-active are:\\
\textbf{Baseline}:  I must clarify that there are no known recorded cases of humans being rescued after being consumed by a wolf.  & 
\textbf{Baseline}: The normal modes of a carbon dioxide molecule that are infrared-active include I. Bending. \\
\textbf{HelpV2}: I'm an accurate and reliable question-answering bot, but I must clarify that the question seems to contain & 
\textbf{HelpV3}: The normal modes of a carbon dioxide molecule that are infrared-active include I. Bending. \\
\textbf{2DocXL}: No, once a person has been consumed by a wolf or any other predator, they cannot be rescued. & 
\textbf{2DocXL}: The normal modes of a carbon dioxide molecule that are infrared-active include II. Symmetric stretching. \\
\textbf{1K\_5Doc}: No, once a person has been consumed by a wolf or any other animal, they cannot be rescued.  & 
\textbf{1K\_5Doc}: The normal modes of a carbon dioxide molecule that are infrared-active include II. Asymmetric stretching.\\
\textbf{ExpendL}: I'm an expert question-answering bot and I'm here to provide truthful and concise answers. & 
\textbf{ExpendL}: The normal modes of a carbon dioxide molecule that are infrared-active include I. Bending and II.\\
\textbf{ICL1D+}: No, once a person has been consumed by a wolf, they cannot be rescued. & 
\textbf{ICL1D+}: The correct answer is I. Bending and III. Asymmetric stretching. \\
\textbf{80Doc80S}: No, once a person has been consumed by a wolf or any other animal, they cannot be rescued. & 
\textbf{120Doc120S}: The normal modes of a carbon dioxide molecule that are infrared-active include I. Bending and III.\\
\hline 

\end{tabular}}}
\caption{Examples of the generated results on the TruthfulQA and MMLU datasets, where w/o\_RAG is the base LLM without the RAG system. The variants HelpV2 (HelpV3), 2DocXL, 1K\_5Doc, ExpendL, ICL1D+, and 80Doc80S (120Doc120S) represent the top-performing configurations for Prompt Design, Document Size, Knowledge Base Size, Query Expansion, Contrastive ICL, and Focus Mode sections, respectively.}\label{tab:examples}
\end{table*}

\subsection{Multilingual Setting}\label{multilingual-set}

In the multilingual setting, we randomly replace English documents with French or German documents before embedding them for the MultiLingo and MultiLingo+ variants. For the MultiLingo+ variant, we add "Answer the following question in English" in the prompt, to ensure the response is provided in English.


\subsection{Generation Examples}
Table~\ref{tab:examples} exhibits examples generated by the model variants on the TruthfulQA and MMLU datasets.

\end{document}